\documentclass[letterpaper, 10 pt, conference]{ieeeconf}  

\IEEEoverridecommandlockouts                              

\usepackage{amsmath} 
\usepackage{amssymb}  
\usepackage{graphicx}
\usepackage{eccvabbrv}
\usepackage{booktabs}
\usepackage{multirow}
\usepackage{xcolor}
\usepackage{bbm}
\usepackage{caption}
\usepackage{subcaption}
\usepackage{hyperref}

\title{\LARGE \bf
VAPO: Visibility-Aware Keypoint Localization\\ 
for Efficient 6DoF Object Pose Estimation
}

\author{Ruyi Lian$^{1}$, Yuewei Lin$^{2}$, Longin Jan Latecki$^{3}$, and Haibin Ling$^{4}$
\thanks{$^{1}$Ruyi Lian is with the Department of Electrical Engineering and Computer Science, South Dakota State University, Brookings, SD 57007 USA (e-mail: Ruyi.Lian@sdstate.edu). This work was conducted in part at Stony Brook University.}
\thanks{$^{2}$Yuewei Lin is with the Computational Science Initiative, Brookhaven National Laboratory, Upton, NY 11973 USA (e-mail: ywlin@bnl.gov).}
\thanks{$^{3}$Longin Jan Latecki is with the Department of Computer and Information Sciences, Temple University, Philadelphia, PA 19122 USA (e-mail: latecki@temple.edu).}
\thanks{$^{4}$Haibin Ling is with the Department of Computer Science, Stony Brook University, New York, NY 11794 USA (e-mail: hling@cs.stonybrook.edu).}
}

\begin{document}

\maketitle
\thispagestyle{empty}
\pagestyle{empty}

\begin{abstract}

Localizing predefined 3D keypoints in a 2D image is an effective way to establish 3D-2D correspondences for instance-level 6DoF object pose estimation. However, unreliable localization results of invisible keypoints degrade the quality of correspondences. In this paper, we address this issue by localizing the important keypoints in terms of visibility. Since keypoint visibility information is currently missing in the dataset collection process, we propose an efficient way to generate binary visibility labels from available object-level annotations, for keypoints of both asymmetric objects and symmetric objects. We further derive real-valued visibility-aware importance from binary labels based on the PageRank algorithm. Taking advantage of the flexibility of our visibility-aware importance, we construct VAPO (Visibility-Aware POse estimator) by integrating the visibility-aware importance with a state-of-the-art pose estimation algorithm, along with additional positional encoding. VAPO can work in both CAD-based and CAD-free settings. Extensive experiments are conducted on popular pose estimation benchmarks including Linemod, Linemod-Occlusion, and YCB-V, 
demonstrating that VAPO clearly achieves state-of-the-art performances.
Project page: \href{https://github.com/RuyiLian/VAPO}{https://github.com/RuyiLian/VAPO}.

\end{abstract}

\section{INTRODUCTION}

Given a single input RGB image, the instance-level 6DoF object pose estimator recovers rotation and translation of a rigid object with respect to a calibrated camera. The pose estimator is crucial in numerous real-world applications, including robot manipulation \cite{araki2021iterative, li20236d, huang2024sdnet}, autonomous driving \cite{tang2023rov6d, rathinam2024spades}, augmented reality \cite{marchand2015pose, tang20193d}, 
\etc. To increase the robustness under various imaging conditions, most existing methods~\cite{rad2017bb8, tekin2018real, oberweger2018making, peng2019pvnet, zakharov2019dpod, park2019pix2pose, li2019cdpn, su2022zebrapose, lian2023checkerpose} first generate correspondences between 2D image pixels and 3D object points, and then regress the pose via any available Perspective-n-Point (PnP) solver~\cite{lepetit2009epnp, wang2021gdr, barath2019progressive}.

Based on the correspondence estimation process, previous methods can be divided into two categories. The first kind of methods~\cite{zakharov2019dpod, park2019pix2pose, li2019cdpn, wang2021gdr, di2021so, su2022zebrapose} estimate corresponding 3D coordinate on the object surface for each 2D pixel.
The second kind of methods~\cite{rad2017bb8, tekin2018real, peng2019pvnet, liu2021kdfnet, lian2023checkerpose} localize predefined 3D keypoints in the input image to obtain 3D-2D correspondences,
which more efficiently encode the object geometry information and facilitate the pose estimation process. 

To obtain better correspondences for object pose estimation, a great amount of effort has been devoted to improve the localization precision of each keypoint. However, existing keypoint-based methods share a common limitation, \textit{i.e.}, a large portion of predefined keypoints are invisible due to occlusion or self-occlusion. Without direct observations, localizing such keypoints often leads to unreliable results. Since the ultimate goal of keypoint localization is to establish reliable 3D-2D correspondences for 6DoF pose estimation, it may not be necessary to localize each predefined keypoint.

To overcome this issue, we propose to estimate visibility-aware importance for each keypoint, and discard unimportant keypoints before localization.
However, annotations of keypoint visibility are currently missing in 6DoF object pose datasets. 
To avoid the expensive manual annotation process, we propose an efficient way to generate visibility labels from available object-level annotations. Visibility is decomposed into two binary terms \wrt~external occlusion and internal self-occlusion. The external visibility term can be obtained from object segmentation masks. The internal visibility term can be determined based on surface normals and camera ray directions, inspired by back-face culling~\cite{kumar1996hierarchical} in rendering.
For symmetric objects, we derive modified computation to ensure consistency. 
We further complement the binary labels with a real-valued measure which is easy to evaluate.
To do this, we create a $k$-nearest neighbor ($k$-NN) graph from the predefined keypoints, and utilize Personalized PageRank (PPR)~\cite{page1998pagerank} to compute the closeness of each keypoint to visible ones as the desired measure.

Our visibility-aware importance can be seamlessly integrated into existing keypoint-based 6DoF pose estimator to boost performance. 
We add a visibility-aware importance predictor before keypoint localization to eliminate keypoints with low importance.
We use positional encoding to enhance the embeddings of the selected keypoints and adopt a two-stage training strategy to efficiently train our pose estimator.
Furthermore, our method can easily adapt to a more general setting where precise 3D CAD model is unavailable.

To summarize, we make the following contributions: 
\begin{itemize}
    \item We propose to localize important keypoints in terms of visibility, to obtain high-quality 3D-2D correspondences for 6DoF object pose estimation. 

    \item From object-level annotations, we derive an efficient way to generate binary keypoint visibility labels and real-valued visibility-aware keypoint importance, for both asymmetric objects and symmetric ones. 


    \item We demonstrate that our visibility-aware importance can be easily incorporated to existing keypoint-based method for both CAD-based and CAD-free settings. 
\end{itemize}
We conduct extensive experiments on Linemod \cite{hinterstoisser2012model}, Linemod-Occlusion \cite{brachmann2014learning}, and YCB-V \cite{xiang2017posecnn} to demonstrate the effectiveness of our method.

\section{Related Work}

In this section, we review previous studies that are most relevant to our work. For a more comprehensive review of 6DoF object pose estimation, we refer readers to~\cite{liu2024deep}.


\noindent\textbf{Correspondence-Based Methods for Instance-level 6DoF Pose Estimation from RGB Inputs.} 
One popular way to establish 3D-2D correspondences can be regarded as an image-to-image translation task~\cite{park2019pix2pose}. Specifically, for each 2D pixel, the corresponding 3D point on the object surface is predicted in the object frame~\cite{park2019pix2pose, li2019cdpn, zakharov2019dpod, wang2021gdr, di2021so, su2022zebrapose}. 
The dense correspondences are robust against various imaging conditions.
Object pose can be recovered from the correspondences via existing PnP solvers~\cite{lepetit2009epnp, barath2019progressive}, coupled with RANSAC to remove outliers. 
Localizing predefined keypoints in the input image is also widely used for constructing 3D-2D correspondences. For simplicity, previous works~\cite{rad2017bb8, tekin2018real, oberweger2018making} localize the 3D object bounding box corners. Other works~\cite{peng2019pvnet, liu2021kdfnet} also adopt sparse keypoints (\eg, 8 keypoints) on the object surface obtained via farthest point sampling (FPS).
The recently proposed CheckerPose~\cite{lian2023checkerpose} localizes dense keypoints to construct dense 3D-2D correspondences, increasing the robustness similar to the image-to-image translation-based methods.  
6D-Diff~\cite{xu20246d} proposes to formulate keypoint localization as a reverse diffusion process. 

\noindent\textbf{CAD-Free Object Pose Estimation.}
To remove the dependence on precise CAD models, CAD-free methods have been studied recently. 
RLLG~\cite{cai2020reconstruct} learns correspondences via multi-view geometric constraints.
OnePose~\cite{sun2022onepose} 
obtains correspondences between the query image and SfM model via a keypoint-based matching network. 
OnePose++~\cite{he2022onepose++} instead uses a keypoint-free feature matching pipeline 
for low-textured objects.
GS-Pose~\cite{cai2024gs} adopts 3D Gaussian Splatting~\cite{kerbl20233d} to build explicit representation of the object. 

\noindent\textbf{Visibility Estimation for 3D Vision and Graphics.} 
Visibility has been used to select reliable correspondences for 6DoF object pose estimation~\cite{lian2023checkerpose} and relative camera pose estimation~\cite{hutchcroft2022covispose}.
Still, each correspondence is generated despite its visibility.
For nonrigid pose estimation, 
keypoint visibility is manually annotated in various datasets~\cite{johnson2010clustered, andriluka20142d, lin2014microsoft} as supervision signals.
In rendering of large polygonal models, back-facing polygons are eliminated to speed up the rendering process~\cite{kumar1996hierarchical}. 

\section{Method}
Our work focuses on instance-level pose estimation for a rigid object $O$.
In the classical setting where the CAD model of $O$ is available,
we sample 3D keypoints $\mathcal{P}$ over the object surface, and estimate the importance of each keypoint $\mathbf{p} \in \mathcal{P}$ \wrt~visibility in the input RGB image $I$.
We then localize the subset of $\mathcal{P}$ with high importance, and obtain rotation $\mathbf{R} \in SO(3)$ and translation $\mathbf{t} \in \mathbb{R}^3$ from the localization results via a PnP solver~\cite{lepetit2009epnp, barath2019progressive}.
Furthermore, our method can be easily extended to the CAD-free setting. 
We describe our method, named VAPO (Visibility-Aware POse estimator), in details as follows.

\subsection{Visibility Labels from Object-Level Annotations}
\label{sec:method_dual_visibility}

\begin{figure}[!t]
  \centering
  \includegraphics[width=0.6\linewidth]{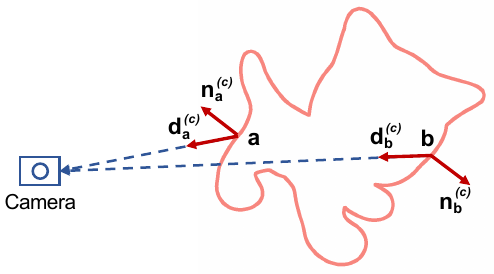}
  \caption{\textbf{Illustration of keypoint internal visibility.} 
  Inspired by back-face culling~\cite{kumar1996hierarchical} in rendering, we compute the internal visibility based on the 
  inner product between the directional vector towards the camera (\eg, $\mathbf{d}_{\mathbf{a}}^{(c)}$) and the normal vector (\eg, $\mathbf{n}_{\mathbf{a}}^{(c)}$).}
  \label{fig:visibility_illustration}
\end{figure}

Keypoint-level annotations are typically unavailable in existing 6DoF object pose datasets~\cite{hinterstoisser2012model, brachmann2014learning, xiang2017posecnn}.
To avoid expensive manual annotation, we seek an efficient way to generate visibility labels from available object-level annotation, \ie, rotation $\mathbf{R}$, translation $\mathbf{t}$, and object segmentation masks $M$.
A keypoint $\mathbf{p} \in \mathcal{P}$ is visible if and only if it is free from both occlusions and self-occlusions. Thus we can decompose visibility into two binary terms: the external visibility $V_{\mathrm{ex}}(\mathbf{p}) \in \{0, 1\}$ \wrt~occlusions from other objects, and the internal visibility $V_{\mathrm{in}}(\mathbf{p}) \in \{0, 1\}$ \wrt~self-occlusions. The overall visibility $V(\mathbf{p}) \in \{0, 1\}$ can be computed by
\begin{equation}
    V(\mathbf{p}) = V_{\mathrm{ex}}(\mathbf{p}) \times V_{\mathrm{in}}(\mathbf{p}),
    \label{eq:visib_overall}
\end{equation}
and keypoint $\mathbf{p}$ satisfies $V(\mathbf{p}) = 1$ if and only if $V_{\mathrm{ex}}(\mathbf{p}) = 1$ and $V_{\mathrm{in}}(\mathbf{p}) = 1$.

Since the visible segmentation mask $M_{\mathrm{vis}}$ of the object $O$ reflects occlusions from other objects, we can determine $V_{\mathrm{ex}}(\mathbf{p})$ by
\begin{equation}
    V_{\mathrm{ex}}(\mathbf{p}) = 
    \begin{cases}
        1, & \text{if } \boldsymbol\pi(\mathbf{p}; \mathbf{R}, \mathbf{t}) \in M_{\mathrm{vis}} \\
        0, & \text{otherwise}
    \end{cases},
    \label{eq:visib_external}
\end{equation}
where $\boldsymbol\pi(\mathbf{p}; \mathbf{R}, \mathbf{t})$ is the perspective projection of $\mathbf{p}$ using pose $(\mathbf{R}, \mathbf{t})$.

To determine whether keypoint $\mathbf{p} \in \mathcal{P}$ is self-occluded, we can check whether the direction from $\mathbf{p}$ towards the camera has additional intersections with the object surface. 
However, it is time-consuming to check the intersections on the fly during training.
Inspired by back-face culling~\cite{kumar1996hierarchical} in rendering, we compute $V_{\mathrm{in}}(\mathbf{p})$ by
\begin{equation}
     V_{\mathrm{in}}(\mathbf{p}) = 
     \begin{cases}
        1, & \text{if } \mathbf{d}_{\mathbf{p}}^{(c)} \cdot \mathbf{n}_{\mathbf{p}}^{(c)} > 0 \\
        0, & \text{otherwise}
    \end{cases},
    \label{eq:visib_inner}
\end{equation}
where $\mathbf{d}_{\mathbf{p}}^{(c)}$ denotes the direction from $\mathbf{p}$ towards the camera in the camera space,  $\mathbf{n}_{\mathbf{p}}^{(c)}$ denotes the surface normal at $\mathbf{p}$ in the camera space, and ``$\cdot$" indicates the dot product.  
As illustrated in Figure~\ref{fig:visibility_illustration}, 
keypoint $\mathbf{a}$ is internally visible while $\mathbf{b}$ is internally invisible according to Eq.~(\ref{eq:visib_inner}).
In the camera space, the camera is placed at the origin $[0, 0, 0]^{\top}$, and then 
\begin{equation}
    \mathbf{d}_{\mathbf{p}}^{(c)} = -\mathbf{p}^{(c)} = - (\mathbf{R} \mathbf{p} + \mathbf{t}), 
    \label{eq:direction_vector}
\end{equation}
where $\mathbf{p}^{(c)}$ is the keypoint coordinate in the camera space. When the 3D CAD model is available, we have access to the surface normal $\mathbf{n}_{\mathbf{p}}$ in the object frame, and we get
\begin{equation}
    \mathbf{n}_{\mathbf{p}}^{(c)} = \mathbf{R} \mathbf{n}_{\mathbf{p}},
    \label{eq:normal_camera_space}
\end{equation}
thus Eq.~(\ref{eq:visib_inner}) can be efficiently computed from available object-level annotations.
When $V_{\mathrm{in}}(\mathbf{p}) = 0$, then $\mathbf{p}$ must be self-occluded, and we can safely eliminate $\mathbf{p}$ or reduce the importance of $\mathbf{p}$ in localization process. When $V_{\mathrm{in}}(\mathbf{p}) = 1$ and the object is non-convex, $\mathbf{d}_{\mathbf{p}}^{(c)}$ may have additional intersections with the object surface, but we omit additional inspections to accelerate the training process. 

\subsection{Handling Visibility of Symmetric Objects}
\label{sec:method_symmetry}


For a symmetric object, an input image corresponds to multiple equivalent poses \wrt~the symmetry transformations $\mathcal{S}$.
When the 3D CAD model is available, $\mathcal{S}$ can be obtained by manual inspection or geometric analysis.
In practice, 
only one of the equivalent poses is annotated.
For different images with similar appearances, the annotated poses may be dramatically different. 
From the perspective of keypoint visibility, it is not optimal to directly use the annotated poses to generate visibility labels, 
because the inconsistency of the labels is problematic for training a robust visibility classifier. 

To enforce the consistency of visibility labels, we can transform the original annotated pose $(\mathbf{R}, \mathbf{t})$ to a canonical one with a proper symmetry transformation $\tilde{\mathbf{S}} \in \mathcal{S}$. 
We select $\tilde{\mathbf{S}}$ by maximizing the number of internally visible keypoints in a fixed subset of keypoints $\mathcal{P}_{\mathrm{sym}}$. We consider internal visibility $V_{\mathrm{in}}$ because self-occlusion is not affected by external objects.

For an object with discrete symmetry, before the whole training process, we obtain $\mathcal{P}_{\mathrm{sym}}$ by finding the largest visible subset under sampled poses. In practice, we uniformly sample 2,562 rotation matrices in $SO(3)$, and use a fixed translation $\mathbf{t}$ along the camera looking direction. 
Then during training, we can enumerate the finite equivalent poses to find the one maximizing internally visible keypoints in $\mathcal{P}_{\mathrm{sym}}$. 

For an object with continuous symmetry, 
we can further derive analytic formulas that are easy to evaluate.
With Eq.~(\ref{eq:direction_vector}) and Eq.~(\ref{eq:normal_camera_space}), we get
\begin{equation}
\begin{split}
    -\mathbf{d}_{\mathbf{p}}^{(c)} \cdot \mathbf{n}_{\mathbf{p}}^{(c)} 
    &= (\mathbf{R} \mathbf{p} + \mathbf{t})^{\top} (\mathbf{R} \mathbf{n}_{\mathbf{p}}) \\
    &= \mathbf{p}^{\top} \mathbf{R}^{\top} \mathbf{R} \mathbf{n}_{\mathbf{p}} + \mathbf{t}^{\top} \mathbf{R} \mathbf{n}_{\mathbf{p}} \\
    &= \mathbf{p}^{\top} \mathbf{n}_{\mathbf{p}} + (\mathbf{R}^{\top} \mathbf{t})^{\top} \mathbf{n}_{\mathbf{p}},
\end{split}
\label{eq:visib_internal_detail}
\end{equation}
and we can see that the first term $\mathbf{p}^{\top} \mathbf{n}_{\mathbf{p}}$ is invariant. To maximize internally visible keypoints in $\mathcal{P}_{\mathrm{sym}}$, we only need to minimize the second term $(\mathbf{R}^{\top} \mathbf{t})^{\top} \mathbf{n}_{\mathbf{p}}$ for $\mathcal{P}_{\mathrm{sym}}$.

\textbf{Base case: single continuous symmetry axis, no additional discrete symmetry.} 
The symmetry transformations can be parameterized as $\mathbf{R}_{*}(\theta)$, where $*$ denotes a symmetry axis and $\theta$ is the rotation angle around the symmetry axis. For a specific transformation $\mathbf{R}_{*}(\theta)$, the corresponding transformed rotation is
\begin{equation}
    \tilde{\mathbf{R}} = \mathbf{R} \mathbf{R}_{*}(\theta),
\end{equation}
and the corresponding transformed translation remains the same, 
thus 
\begin{equation}
    (\tilde{\mathbf{R}}^{\top} \tilde{\mathbf{t}})^{\top} \mathbf{n}_{\mathbf{p}} 
    = (\mathbf{R}^{\top} \mathbf{t})^{\top} (\mathbf{R}_{*}(\theta) \mathbf{n}_{\mathbf{p}}).
\end{equation}

Without loss of generality, we can assume $z$-axis is the symmetry axis. Otherwise, we can apply a simple coordinate system transformation to the object to satisfy this assumption. Consider a special case when $\mathbf{n}_{\mathbf{p}} = \mathbf{n}_{0} = [1, 0, 0]^{\top}$, and then
\begin{equation}
    \mathbf{R}_{z}(\theta) \mathbf{n}_{0} = 
    \begin{bmatrix}
        \cos{\theta} & -\sin{\theta} & 0 \\
        \sin{\theta} & \cos{\theta} & 0 \\
        0 & 0 & 1 
    \end{bmatrix}
    \begin{bmatrix}
        1 \\
        0 \\
        0
    \end{bmatrix}
    = \begin{bmatrix}
        \cos{\theta} \\
        \sin{\theta} \\
        0
    \end{bmatrix},
\end{equation}
and we further get
\begin{equation}
    (\tilde{\mathbf{R}}^{\top} \tilde{\mathbf{t}})^{\top} \mathbf{n}_{0} 
    = a \cos{\theta} + b \sin{\theta}, 
\end{equation}
where $a, b$ are the first and second entry of $\mathbf{R}^{\top} \mathbf{t}$, respectively. 
For simplicity, denote $(\tilde{\mathbf{R}}^{\top} \tilde{\mathbf{t}})^{\top} \mathbf{n}_{0}$ as $f(\theta)$,
and now we need to minimize $f(\theta)$. When $a = 0$, it is straightforward to get 
$\theta = \pi / 2 + [b > 0] \pi$, where $[\cdot]$ is Iverson bracket outputting binary values $\{0, 1\}$. When $a \neq 0$, we have
\begin{equation}
    f^{\prime}(\theta) = -a \sin\theta + b\cos\theta,
\end{equation}
and $\theta = \arctan{(b / a)}$ is one of the solutions to make $f^{\prime}(\theta) = 0$. We also need to consider 
\begin{equation}
    f^{\prime\prime}(\theta) = -a\cos\theta -b\sin\theta 
    = -a\cos\theta(1 + \frac{b^2}{a^2}),
\end{equation}
and let $f^{\prime\prime}(\theta) \geq 0$. 
Thus when 
\begin{equation}
    \theta = 
     \begin{cases}
        \pi / 2 + [b > 0] \pi, & \text{if } a = 0 \\
        \arctan{(b / a)} + [a > 0] \pi, & \text{otherwise }
    \end{cases},
    \label{eq:cont_sym_analytic}
\end{equation}
$(\tilde{\mathbf{R}}^{\top} \tilde{\mathbf{t}})^{\top} \mathbf{n}_{0}$ achieves its minimum value, thus the quantity in Eq.~(\ref{eq:visib_internal_detail}) also achives its minimum value 
for the point $\mathbf{p}_0$ with $\mathbf{n}_{\mathbf{p}} = \mathbf{n}_{0} = [1, 0, 0]^{\top}$ and its neighborhood, which ensures the internal visibility of $\mathbf{p}_0$ and its neighborhood (\ie, $\mathcal{P}_{\mathrm{sym}}$). 


In summary, we modify the annotated rotation matrix by right-multiplying rotation around $z$-axis $\mathbf{R}_z(\theta)$, where $\theta$ is the rotation angle and is determined by Eq.~(\ref{eq:cont_sym_analytic}).

\textbf{Variation I: single continuous symmetry axis, additional discrete symmetry.} 
We can first 
Eq.~(\ref{eq:cont_sym_analytic}) to resolve the ambiguity caused by continuous symmetry, then enumerate the finite equivalent poses of the discrete symmetry.

\textbf{Variation II: multiple continuous symmetry axes.} 
Without loss of generality, we can assume $y$-axis is another symmetry axis. After resolving the ambiguity caused by continuous symmetry of $z$-axis, we apply additional transformation $\mathbf{R}_{y}(\phi)$, where $\phi$ is the rotation angle around $y$-axis. The derivation of $\phi$ is very similar to $\theta$ in $\mathbf{R}_z(\theta)$, hence we omit it for simplicity.

\begin{figure*}[!t]
  \centering
  \includegraphics[width=0.9\linewidth]{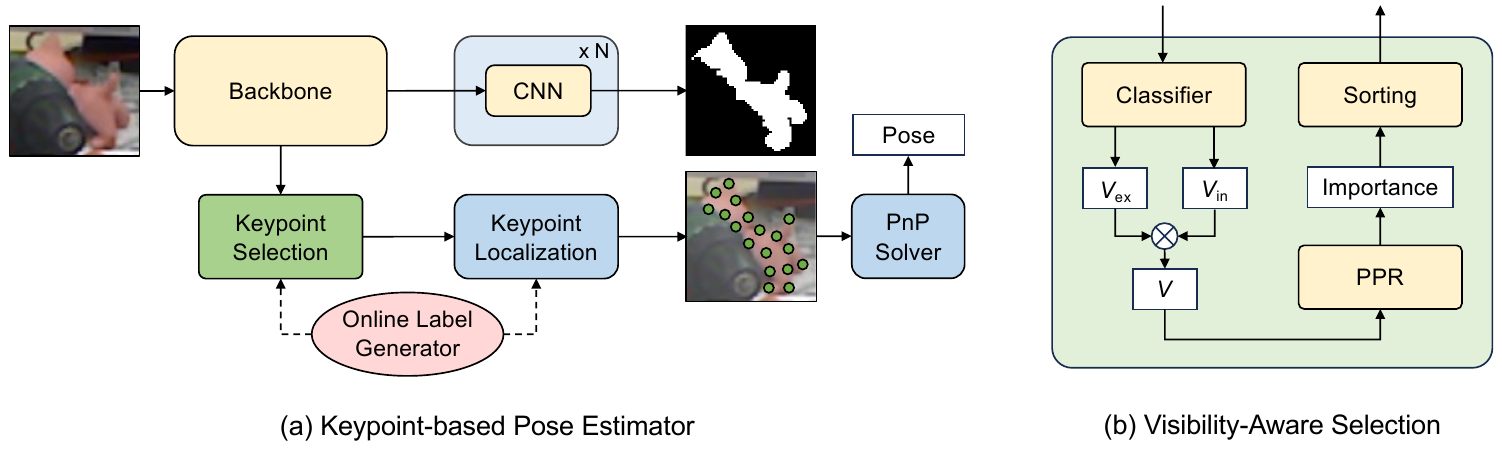}
  \caption{\textbf{Illustration of our visibility-aware pose estimator.}
  (a) Overall architecture: We use a backbone network to extract features from the input image, and select keypoints with high visibility-aware importance for localization. 
  We generate supervision signals online for keypoint visibility and localization. 
  We also train object segmentation with available ground truth masks. 
  We finally recover the pose using a PnP solver.
  (b) Details of visibility-aware selection module: For each keypoint, we use a multi-label classifier to predict external visibility $V_{\mathrm{ex}}$ and internal visibility $V_{\mathrm{in}}$. We then obtain overall visibility $V$, and adopt Personalized PageRank (PPR) to generate visibility-aware importance from $V$. 
  }
  \label{fig:pipeline}
\end{figure*}

\subsection{Visibility-Aware Importance via Personalized PageRank}
\label{sec:method_ppr}

We can further compute real-valued visibility-aware importance by measuring the closeness to visible keypoints \wrt~a specific measure of the proximity.
In practice, we adopt Personalized PageRank (PPR) as our proximity measure, which is a variation of PageRank~\cite{page1998pagerank} based on a random walk model. 
Specifically, we build a directed $k$-nearest neighbor ($k$-NN) graph $G$ from predefined keypoints $\mathcal{P}$. For each keypoint $\mathbf{p}$, we create edges from $\mathbf{p}$ to its $k$-nearest neighbors. We define transition matrix $\mathbf{T}$ as
\begin{equation}
    \mathbf{T} = \frac{1}{k} \mathbf{A}^{\top},
    \label{eq:def_transition_matrix}
\end{equation}
where $\mathbf{A}$ is the adjacency matrix of $G$. With probability $c \in (0, 1)$, a random walker on $G$ moves along edges following $\mathbf{T}^{\top}$. 
With probability $1 - c$, the random walker restarts at any visible keypoint with uniform probability. $c$ is often called damping factor, and we assign $c$ the widely used value 0.85.
We use the stationary probability distribution $\mathbf{r}$ over $N$ keypoints to measure closeness to visible keypoints, where each entry of $\mathbf{r}$ represents the probability that the random walker resides on the corresponding keypoint.
We can obtain $\mathbf{r}$ by solving the following equation
\begin{equation}
    \mathbf{r} = c \mathbf{T} \mathbf{r} + (1-c) \mathbf{s},
    \label{eq:ppr_random_walk}
\end{equation}
where $\mathbf{s}$ is the restart vector, and the entry of keypoint $\mathbf{p}$ is determined from binary visibility labels as 
\begin{equation}
     \mathbf{s}(\mathbf{p}) = 
     \begin{cases}
        1 / N_{\mathrm{vis}}, & \text{if } V(\mathbf{p}) = 1 \\
        0, & \text{otherwise}
    \end{cases},
    \label{eq:def_restart_vector}
\end{equation}
where $N_{\mathrm{vis}}$ denotes the number of visible keypoints.
By rearranging Eq.~(\ref{eq:ppr_random_walk}), we can compute $\mathbf{r}$ as
\begin{equation}
    \mathbf{r} = \mathbf{T}_{\mathrm{ppr}} \mathbf{s},
    \label{eq:ppr_solve}
\end{equation}
where 
\begin{equation}
    \mathbf{T}_{\mathrm{ppr}} \stackrel{\text{def}}{=} (1-c) (\mathbf{I} - c\mathbf{T})^{-1}.
    \label{eq:ppr_precompute_coef}
\end{equation}
Note that $\mathbf{T}_{\mathrm{ppr}}$ is well-defined in Eq.~(\ref{eq:ppr_precompute_coef}) because $\mathbf{I} - c\mathbf{T}$ is invertible.
Moreover, $\mathbf{T}_{\mathrm{ppr}}$ is invariant for the object, so we can precompute $\mathbf{T}_{\mathrm{ppr}}$ using Eq.~(\ref{eq:ppr_precompute_coef}) and store it.


\subsection{From Visibility-Aware Importance to Pose Estimation}
\label{sec:method_checker}

We first use binary classifiers to predict binary visibility $V$ (Section~\ref{sec:method_dual_visibility}) and compute restart vector $\mathbf{s}$ (Eq.~(\ref{eq:def_restart_vector})), then we get importance $\mathbf{r}$ using Eq.~(\ref{eq:ppr_solve}). In this way, we only need to train binary classifiers, which is typically easier than training regression models of $\mathbf{r}$. 
Our proposed visibility-aware importance $\mathbf{r}$ can be easily integrated into existing keypoint-based 6DoF pose estimators. 
We adopt the recent state-of-the-art open-source method CheckerPose~\cite{lian2023checkerpose} as our base.
CheckerPose utilizes GNNs of dense keypoints and predicts binary codes as a hierarchical representation of 2D locations.
The 2D locations can be easily obtained from this representation without voting schemes.
Besides, CheckerPose exploits a CNN decoder to learn image features and fuses the features into the GNN branch.
To facilitate mini-batch-based training for GNNs, we select $N^{\prime} = N/2$ keypoints with highest importance for localization. 
For extreme case when the ratio of estimated visible keypoints is below a certain threshold, we directly use $N^{\prime}$ evenly distributed keypoints for robustness.

As shown in Figure~\ref{fig:pipeline}, we use a backbone network to extract image features from input region of interest (RoI) $I_O$ of object $O$. Then we utilize a multi-label classifier to predict external visibility $V_{\mathrm{ex}}(\mathbf{p})$ and internal visibility $V_{\mathrm{in}}(\mathbf{p})$ for each keypoint $\mathbf{p} \in \mathcal{P}$. 
Within the classifier, we adopt a shallow GNN on $k$-NN graph $G$ described in Section~\ref{sec:method_ppr}.
We can get overall binary visibility $V(\mathbf{p})$ by simply multiplying $V_{\mathrm{ex}}(\mathbf{p})$ and $V_{\mathrm{in}}(\mathbf{p})$,
and generate real-valued visibility-aware importance $\mathbf{r}$ using PPR based algorithm (Eq.~(\ref{eq:ppr_solve})). 
We then select $N^{\prime} = N/2$ keypoints with highest importance, denoted as $\mathcal{P}_{\mathrm{vis}} \subset \mathcal{P}$. We use the subgraph $G_{\mathrm{vis}}$ of $G$ induced by $\mathcal{P}_{\mathrm{vis}}$ in GNN-based localization process.

We further enhance keypoint embedding to improve the network performance. Specifically, We use a shallow GNN to obtain $64$-dim embedding $F_P$ from the coordinates of $\mathcal{P}$, and concatenate it with initial keypoint embedding in $G$. We can regard $F_P$ as positional encoding~\cite{mildenhall2020nerf}, which facilitates the localization of the dynamic subgraph $G_{\mathrm{vis}}$.

We use a two-stage training procedure to train our visibility-aware pose estimator. 
At the first stage, we train network layers corresponding to low-level estimations, including external visibility $V_{\mathrm{ex}}$, internal visibility $V_{\mathrm{in}}$, 1-bit indicator code $\mathbf{b_v}$, and the first $d_0$ bits of $\mathbf{b_x}, \mathbf{b_y}$ ($\mathbf{b_v},\mathbf{b_x}$ and $\mathbf{b_y}$ are defined in CheckerPose).
At the second stage, we train the whole network. 
We use binary cross-entropy loss for visibility estimation and binary code generation, and apply $L_1$ loss for segmentation mask prediction. 

\subsection{Extension for CAD-Free Object Pose Estimation}
\label{sec:method_cad_free}
When the 3D CAD model is unavailable, 
we can use an off-the-shelf image-based mesh reconstructor.
The biggest problem is the imperfect reconstruction quality,
especially the inaccurate reconstruction of invisible regions.
To address this issue, for each object, we use $m$ views to reconstruct $m$ 3D meshes.
For each reconstructed mesh,
we use the criteria in Section~\ref{sec:method_dual_visibility} to determine the visible vertices in the corresponding input view, from which we uniformly sample $N/m$ keypoints. In this way, we obtain predefined $N$ keypoints with plausible 3D coordinates and normals. 
Furthermore, when generating synthetic training images, for each randomly sampled pose, we render the mesh reconstructed from the closest view. 
Once we 
generate sufficient training images, we can train VAPO following the CAD-based setting.
When prior knowledge of symmetry is unavailable, we assume that all objects are asymmetric during training.

\section{Experiments}

\subsection{Experimental Setup}

\textbf{Datasets.} Following the common practice, we conduct extensive experiments on Linemod (LM) \cite{hinterstoisser2012model}, Linemod-Occlusion (LM-O) \cite{brachmann2014learning}, and YCB-V \cite{xiang2017posecnn}. 
LM contains 13 objects, and provides around $1,200$ real images for each object with mild occlusions. Following \cite{brachmann2016uncertainty}, we use about $15\%$ images as training set and test our method on the remaining images. We also use $1,000$ synthetic training images for each object following \cite{li2019cdpn, wang2021gdr, di2021so}. 
LM-O contains 8 objects from LM, and uses the real images from LM as training images. The test set is composed of $1,214$ real images with severe occlusions and clutters. 
YCB-V contains 21 daily objects and provides more than $110,000$ real images with severe occlusions and clutters. For training on LM-O and YCB-V, we follow the recent trend~\cite{wang2021gdr, su2022zebrapose, lian2023checkerpose} and use the physically-based rendered data \cite{hodan2020bop} as additional training images.

\textbf{Implementation Details.} We implement our method using PyTorch~\cite{paszke2019pytorch}. 
We train our network using the Adam optimizer~\cite{kingma2014adam} with a batch size of 32. 
Following~\cite{lian2023checkerpose}, we train a unified pose estimator for all 13 LM objects, while training separate pose estimators for each object on LM-O and YCB-V.
The training follows a two-stage procedure. 
In the first stage, we train the layers generating low-level quantities for 50k steps with a learning rate of 2e-4. 
In the second stage, for LM, we continue training all network layers for 100k steps, followed by an additional 20k steps with a reduced learning rate of 1e-4. 
We train all network layers for 700k steps on LM-O and 250k steps on YCB-V, both using a learning rate of 2e-4.
We adopt CheckerPose~\cite{lian2023checkerpose} as our base keypoint localization method, and follow it to use $N = 512$ predefined keypoints and $k = 20$ nearest neighbors for $k$-NN graph. We select $N^{\prime} = 256$ keypoints and use the induced subgraph in the localization step. For the binary code generation, we set the number of bits $d$ as 7, and the number of initial bits $d_0$ as 3. 
We apply data augmentation and off-the-shelf object detectors following~\cite{wang2021gdr, lian2023checkerpose}.
To obtain final poses, we run Progressive-X for 400 iterations with a reprojection error threshold of 2 pixels.
For the CAD-free setting, we use Wonder3D~\cite{long2024wonder3d} to reconstruct $m=8$ meshes from $m=8$ views.

\textbf{Evaluation Metrics.} We adopt the commonly-used metric ADD(-S) to evaluate the estimated poses. To compute ADD(-S) with threshold $x\%$, we transform the 3D model points using the predicted poses and the ground truth, compute the average distance between the transformed results, and check whether the average distance is below $x\%$ of the object diameter. For symmetric objects, we compute the average distance based on the closest points. 
On YCB-V, we also
report the AUC (area under curve) of ADD-S and ADD(-S)~\cite{xiang2017posecnn}, where the symmetric metric is used for all objects in ADD-S but for symmetric objects only in ADD(-S). 

\subsection{Ablation Study on LM Dataset}

\begin{table}[tb]
  \caption{\textbf{Ablation Study on LM.} We highlight the best result and the second best result in red and blue, respectively. 
  }
  \centering
  \small
  \renewcommand{\tabcolsep}{1.5mm}
\begin{tabular}{ c | c | c | c | c }
\toprule
\multirow{2}{*}{Method} & \multicolumn{3}{c|}{ADD(-S)} & \multirow{2}{*}{MEAN} \\
\cline{2-4}
 & 0.02d  &  0.05d  &  0.1d  &   \\
\midrule
GDR-Net~\cite{wang2021gdr}   & 35.5 & 76.3 & 93.7 & 68.5 \\
SO-Pose~\cite{di2021so}      & 45.9	& 83.1 & 96.0 & 75.0 \\
EPro-PnP~\cite{chen2022epro} & 44.8 & 82.0 & 95.8 & 74.2 \\
CheckerPose~\cite{lian2023checkerpose} & 35.7 & 84.5 & \textcolor{red}{97.1} & 72.4 \\
\midrule
Ours (w/o Selection, $N=256)$ & 44.7 & 84.3 & 96.9 & 75.3 \\
Ours (w/o Selection, $N=512)$ & 45.5 & 84.0 & 96.4 & 75.3 \\ 
\midrule
Ours (w/o $V_{\mathrm{ex}}$) & {48.1} & 85.6 & 96.9 & 76.9 \\
Ours (w/o $V_{\mathrm{in}}$)  & 36.0 & 76.0 & 93.9 & 68.6 \\
Ours (w/o P.~E.) & 46.6 & 84.4 & 96.6 & 75.9 \\
Ours (w/o Two-stage) & 45.4 & 83.9 & 96.1 & 75.1 \\
\midrule
Ours ($N^{\prime} = 128$) & 37.6 & 79.4 & 95.6 & 70.9 \\
Ours ($N^{\prime} = 192$) & 45.4 & 84.6 & 96.8 & 75.6 \\
Ours ($N^{\prime} = 320$) & \textcolor{blue}{48.6} & {85.8} &  \textcolor{blue}{97.0} &  {77.1} \\
\midrule
Ours ($k = 15$) & 48.0  & \textcolor{red}{86.1} & \textcolor{red}{97.1} & 77.1 \\
\midrule
Ours ($c=0.9$) & 47.6 & 85.5 & \textcolor{blue}{97.0} & 76.7 \\
Ours ($c=0.85$) & \textcolor{blue}{48.6} & \textcolor{blue}{85.9} &  \textcolor{blue}{97.0} & \textcolor{blue}{77.2} \\
Ours ($c = 0.8$) & \textcolor{red}{49.2} & \textcolor{red}{86.1} & \textcolor{blue}{97.0} & \textcolor{red}{77.4} \\
\bottomrule
\end{tabular}
  \label{tab:lm_ablation}
\end{table}

\begin{table}[tb]
  \caption{\textbf{Comparison on LM-O.} We report the Average Recall (\%) of ADD(-S) with three thresholds: 0.02d, 0.05d, and 0.1d. 
  We highlight the best result and the second best result in red and blue, respectively. ``--" denotes unavailable results due to the absence of models for evaluation with the specified threshold.
  }
  \centering
  \small
  \renewcommand{\tabcolsep}{1.2mm}
\begin{tabular}{c|c|c|c|c|c|c}
\toprule
\multirow{2}{*}{Method} & GDR & Zebra & LC & Checker & 6D-Diff & \multirow{2}{*}{Ours} \\
 & \cite{wang2021gdr} & \cite{su2022zebrapose} & \cite{liu2023linear} & \cite{lian2023checkerpose} & \cite{xu20246d} & \\
\midrule
0.02d  & 4.4  & \textcolor{red}{9.8}  & 8.6   & 7.3 & --  & \textcolor{blue}{9.7} \\
0.05d  & 31.1 & \textcolor{blue}{44.6} & 44.2  & 43.5 & -- & \textcolor{red}{46.2} \\
0.1d   & 62.2 & 76.9 & \textcolor{blue}{78.06} & 77.5 & \textcolor{red}{79.6} & 78.02 \\
\midrule
Mean   & 32.6 & \textcolor{blue}{43.8} & 43.6 & 42.8 & -- & \textcolor{red}{44.6} \\ 
\bottomrule
\end{tabular}
\label{tab:result_lmo}
\end{table}

\begin{table}[tb]
  \caption{\textbf{Comparison on YCB-V.} We report the average ADD(-S), and AUC of ADD-S and ADD(-S). AUC values computed without and with 11-point interpolation are denoted as ``w/o" and ``w/ IT", respectively. We highlight the best result and the second best result in red and blue, respectively. ``--" denotes unavailable results.}
  \centering
  \small
  \renewcommand{\tabcolsep}{1.3mm}
\begin{tabular}{l|c|c|c|c|c}
\toprule
\multirow{2}{*}{Method} & \multirow{2}{*}{ADD(-S)} & \multicolumn{2}{c|}{AUC-S} & \multicolumn{2}{c}{AUC(-S)} \\
\cline{3-6}
 & & w/o & w/ IT &  w/o & w/ IT \\
\midrule
GDR-Net~\cite{wang2021gdr}          & 60.1 & --   & 91.6 & -- & 84.4 \\
SO-Pose~\cite{di2021so}             & 56.8 & -- & 90.9 & -- & 83.9 \\
Zebra~\cite{su2022zebrapose}        & 80.5 & 90.1 & -- & 85.3 & -- \\
Checker~\cite{lian2023checkerpose}  & 81.4 & 91.3 & \textcolor{blue}{95.3} & 86.4 & \textcolor{blue}{91.1} \\
Zebra-LC~\cite{liu2023linear}       & 82.4 & 90.8 & 95.0 & 86.1 & 90.8 \\
6D-Diff~\cite{xu20246d}             & \textcolor{blue}{83.8} & \textcolor{blue}{91.5} & -- & \textcolor{blue}{87.0} & -- \\
\midrule
Ours & \textcolor{red}{84.9} & \textcolor{red}{92.3} & \textcolor{red}{96.4} & \textcolor{red}{87.9} & \textcolor{red}{92.7} \\
\bottomrule
\end{tabular}
  \label{tab:result_ycbv}
\end{table}

\begin{table}[]
    \caption{\textbf{Reconstruction error on LM}. We use Chamfer distance to evaluate single-view reconstruction and our proposed combination (Section~\ref{sec:method_cad_free}). All results are percentage of the object's diameter for scale-invariant comparison.}
    \centering
    \small
    \renewcommand{\tabcolsep}{1.2mm}
    \begin{tabular}{c|c|c|c|c|c|c|c|c|c}
\toprule
View & 1 & 2 & 3 & 4 & 5 & 6 & 7 & 8 & Ours \\
\midrule
ape & 17.0 & 17.7 & 18.4 & 19.5 & 26.6 & 10.0 & 24.8 & 20.3 & \textbf{9.7} \\ 
benchv. & 21.9 & 17.3 & 14.4 & 12.2 & 20.0 & \textbf{6.8} & 20.2 & 13.9 & 10.9 \\ 
camera & 13.5 & 14.9 & 18.4 & 12.8 & \textbf{10.8} & 13.4 & 19.0 & 17.6 & 11.5 \\ 
can & 13.4 & 19.9 & \textbf{8.1} & 12.2 & 11.8 & 14.9 & 8.6 & 15.7 & 9.6 \\ 
cat & 17.3 & 21.6 & 9.5 & 13.9 & 15.5 & 17.9 & 15.9 & 8.7 & \textbf{8.2} \\ 
driller & 27.0 & 21.5 & 21.4 & 13.0 & 17.7 & 24.6 & 26.7 & \textbf{10.9} & 12.5 \\ 
duck & 18.4 & 14.4 & 15.7 & 10.0 & \textbf{8.4} & 11.3 & 20.8 & 13.5 & 10.1 \\ 
eggbox & 8.7 & 13.2 & 8.7 & 11.6 & 12.5 & 15.6 & 15.6 & 11.3 & \textbf{7.5} \\ 
glue & 19.9 & 14.4 & 9.8 & \textbf{8.8} & 9.5 & 13.8 & 13.7 & 14.7 & 10.1 \\ 
holep. & 21.4 & 12.6 & 19.5 & 17.2 & 15.5 & 19.1 & 13.8 & 19.5 & \textbf{9.5} \\ 
iron & 13.3 & 18.4 & 12.0 & 23.4 & 13.5 & 13.9 & 20.6 & 15.7 & \textbf{9.9} \\ 
lamp & 16.5 & 21.5 & 11.9 & 9.0 & 15.7 & \textbf{8.3} & 22.6 & 11.6 & 9.7 \\ 
phone & 24.8 & 26.8 & \textbf{7.7} & 18.0 & 20.8 & 15.0 & 24.0 & 14.6 & 14.4 \\ 
\bottomrule
    \end{tabular}
    \label{tab:eval_wonder3d_recon}
\end{table}

\begin{table}
    \caption{\textbf{CAD-Free Object Pose Estimation on LM.} We report ADD(-S) (0.1d) and highlight the best result and the second best result in red and blue, respectively.}
    \centering
    \small
    \renewcommand{\tabcolsep}{1.0mm}
    \begin{tabular}{c|c|c|c|c|c}
    \toprule
       Method  & RLLG~\cite{cai2020reconstruct} & One~\cite{sun2022onepose} & One++~\cite{he2022onepose++} & GS~\cite{cai2024gs} & \textbf{Ours} \\
    \midrule
ape & 52.9 & 11.8 & 31.2 & \textcolor{blue}{71.0} & \textcolor{red}{83.4} \\
benchv. & 96.5 & 92.6 & 97.3 & \textcolor{red}{99.8} & \textcolor{blue}{98.9} \\
camera & 87.8 & 88.1 & 88.0 & \textcolor{red}{98.2} & \textcolor{blue}{95.9} \\
can & 86.8 & 77.2 & 89.8 & \textcolor{blue}{97.7} & \textcolor{red}{99.4} \\
cat & 67.3 & 47.9 & 70.4 & \textcolor{blue}{86.7} & \textcolor{red}{94.7} \\
driller & 88.7 & 74.5 & 92.5 & \textcolor{blue}{96.2} & \textcolor{red}{96.7} \\
duck & 54.7 & 34.2 & 42.3 & \textcolor{blue}{77.2} & \textcolor{red}{82.4} \\
eggbox & 94.7 & 71.3 & \textcolor{red}{99.7} & \textcolor{blue}{99.6} & \textcolor{blue}{99.6} \\
glue & 91.9 & 37.5 & 48.0 & \textcolor{blue}{98.4} & \textcolor{red}{99.4} \\
holep. & 75.4 & 54.9 & 69.7 & \textcolor{blue}{87.4} & \textcolor{red}{90.9} \\
iron & 94.5 & 89.2 & \textcolor{blue}{97.4} & \textcolor{red}{99.2} & 95.3 \\
lamp & 96.6 & 87.6 & 97.8 & \textcolor{blue}{98.9} & \textcolor{red}{99.1} \\
phone & \textcolor{blue}{89.2} & 60.6 & 76.0 & 85.0 & \textcolor{red}{93.7} \\
\midrule
mean & 82.9 & 63.6 & 76.9 & \textcolor{blue}{92.0} & \textcolor{red}{94.6} \\
    \bottomrule
    \end{tabular}
    \label{tab:cad_free_lm}
\end{table}


\textbf{Effectiveness of Visibility-Aware Keypoint Selection.} 
We report the performance of using a fixed set of $N$ evenly distributed keypoints in localization step in Table~\ref{tab:lm_ablation}. The results (denoted as w/o Selection) with $N=256$ and $N=512$ degrade especially for ADD(-S) 0.02d, which clearly demonstrates the effectiveness of our visibility-aware keypoint selection scheme. 

\textbf{Effectiveness of Dual Visibility Estimation.} Our methods select keypoints based on external visibility $V_{\mathrm{ex}}$ and internal visibility $V_{\mathrm{in}}$. 
Since the external occlusion is mild in LM, 
selecting keypoints based on $V_{\mathrm{in}}$ (denoted as w/o $V_{\mathrm{ex}}$) can significantly boost the performance compared with w/o Selection. Incorporating $V_{\mathrm{ex}}$ further improves the performance.  

\textbf{Effectiveness of Positional Encoding.} 
Since the selected keypoints are dynamic \wrt~input images, we find that adding positional encoding (denoted as w/o P.~E.) greatly improves the performance. 

\textbf{Effectiveness of Two-stage Training.} In Table~\ref{tab:lm_ablation}, we also report the performance without two-stage training (denoted as w/o Two-stage). The overall performance degrades without two-stage training, since the induced subgraph $G_{\mathrm{vis}}$ is not stable when the visibility estimator does not converge.

\textbf{Number of Selected Keypoints.} We show the results with different number of selected keypoints ($N^{\prime}$) in Table~\ref{tab:lm_ablation}. The performance improves when $N^{\prime}$ is gradually increased towards 256, and remains almost the same when $N^{\prime}$ is increased to 320 with more computational cost. 

\textbf{Choice of Hyperparameters.}
Due to computational constraints, we use $k=20$ to construct the $k$-NN graph. We also evaluate $k=15$, which achieves comparable results as shown in Table~\ref{tab:lm_ablation}. 
For the damping factor $c$, in addition to the commonly used value of 0.85, we also evaluate $c=0.9$ and $c=0.8$ to assess sensitivity in Table~\ref{tab:lm_ablation}. The ADD(-S) (0.1d) metric remains unchanged for different $c$, while ADD(-S) (0.02d) and ADD(-S) (0.05d) improve as $c$ decreases.

\begin{figure*}[!t]
  \centering
  \begin{subfigure}[b]{0.15\linewidth}
    \centering
    \includegraphics[width=\linewidth]{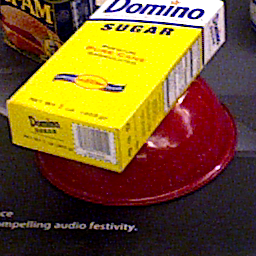}
    \caption{Input Image}
  \end{subfigure}
  \hfill
  \begin{subfigure}[b]{0.15\linewidth}
    \centering
    \includegraphics[width=\linewidth]{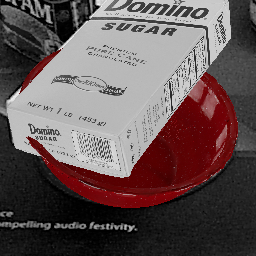}
    \caption{GDR-Net~\cite{wang2021gdr}}
  \end{subfigure}
  \hfill
  \begin{subfigure}[b]{0.15\linewidth}
    \centering
    \includegraphics[width=\linewidth]{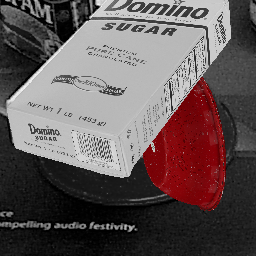}
    \caption{ZebraPose~\cite{su2022zebrapose}}
  \end{subfigure}
  \hfill
  \begin{subfigure}[b]{0.15\linewidth}
    \centering
    \includegraphics[width=\linewidth]{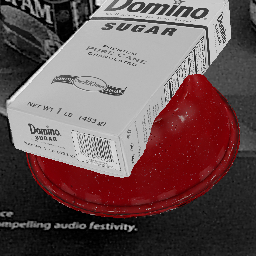}
    \caption{Checker~\cite{lian2023checkerpose}}
  \end{subfigure}
  \hfill
  \begin{subfigure}[b]{0.15\linewidth}
    \centering
    \includegraphics[width=\linewidth]{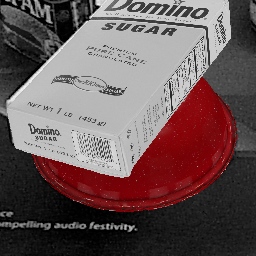}
    \caption{Ours}
  \end{subfigure}
  \hfill
  \begin{subfigure}[b]{0.15\linewidth}
    \centering
    \includegraphics[width=\linewidth]{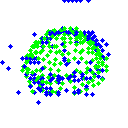}
    \caption{Keypoints}
  \end{subfigure}
\caption{\textbf{Qualitative results}.
In the input image (a), the red bowl is flipped and severely occluded.
In (b)-(e), we render the bowl based on the predictions of different methods.
Background pixels are changed to grayscale values for better visualization. 
In (f) we visualize the localization results of the selected keypoints, generated by CheckerPose (blue) and our method (green).
}
\label{fig:qual_bowl}
\end{figure*}

\subsection{Comparison to State of the Art}

\textbf{Experiments on LM.} As shown in Table~\ref{tab:lm_ablation}, our method significantly improves the performance \wrt~ADD(-S) with threshold $2\%$ and $5\%$ (denoted as 0.02d and 0.05d). This demonstrates that our visibility-aware framework can greatly increase the ratio of qualified poses \wrt~a strict threshold. Our method achieves comparable results \wrt~ADD(-S) with threshold $10\%$ (denoted as 0.1d). The average of the metrics is the best among all methods.

\textbf{Experiments on LM-O.} 
We report the average recall of ADD(-S) metric with three thresholds (0.02d, 0.05d, and 0.1d) in Table~\ref{tab:result_lmo}. As shown in Table~\ref{tab:result_lmo}, our visibility-aware keypoint localization scheme clearly boosts our base CheckerPose~\cite{lian2023checkerpose} on all three thresholds. Our method also greatly surpasses previous methods \wrt~0.05d threshold and the mean of three thresholds, and achieves comparable performance \wrt~other thresholds.

\textbf{Experiments on YCB-V.} 
The real-world test set of YCB-V includes diverse objects in novel and cluttered scenes with severe occlusion, which can evaluate generalizability in realistic settings.
We report the average values of ADD(-S) (0.1d) and AUC in Table~\ref{tab:result_ycbv}.
Our method significantly improves the pose estimation performance \wrt~the ADD(-S).
Our method also achieves the best performance \wrt~ AUC-based metrics, which indicates that our method achieves the best accumulated performance across various thresholds.

\subsection{CAD-Free Object Pose Estimation on LM Dataset}

\begin{table}
    \caption{\textbf{Runtime Analysis.} We highlight the best result and the second best result in red and blue, respectively.}
    \centering
    \small
    \renewcommand{\tabcolsep}{1.0mm}
    \begin{tabular}{c|c|c|c}
    \toprule
       Method  & Corr. & PnP & Overall \\
    \midrule
       Zebra~\cite{su2022zebrapose}  & \textcolor{red}{13.6} & 304.2 & 317.8 \\
       Checker~\cite{lian2023checkerpose} & 68.4 & \textcolor{blue}{33.4} & \textcolor{blue}{101.8} \\
       \midrule
       Ours & \textcolor{blue}{68.3} & \textcolor{red}{31.3} & \textcolor{red}{99.6} \\
    \bottomrule
    \end{tabular}
    \label{tab:runtime}
\end{table}

\begin{table}
    \caption{\textbf{Analysis on the inner visibility computed by Eq.~\ref{eq:visib_inner} on LM objects.} ``Binary" denotes the accuracy of the binary visibility labels, while ``PPR" denotes the accuracy of the keypoints with top PPR-based importance.}
    \centering
    \small
    \renewcommand{\tabcolsep}{1.0mm}
    \begin{tabular}{ccc|ccc|ccc}
    \toprule
    Object & Binary & PPR & Object & Binary & PPR & Object & Binary & PPR \\
    \midrule
    ape & 87.6 & 93.4 & benchv. & 66.0 & 82.8 & camera & 83.5 & 92.0 \\
    can & 83.5 & 93.0 & cat & 80.6 & 88.0 & driller & 86.8 & 89.5 \\
    duck & 89.0 & 91.9 & eggbox & 82.0 & 91.6 & glue & 94.4 & 96.9 \\
    holep. & 60.4 & 67.3 & iron & 87.7 & 91.9 & lamp & 80.4 & 84.5 \\
      &  &  & phone & 83.2 & 91.4 &  &  &  \\   
    \bottomrule
    \end{tabular}
    \label{tab:non_convex}
\end{table}

We evaluate our method in the CAD-free setting on the LM dataset. As shown in Table~\ref{tab:eval_wonder3d_recon}, our strategy of combining visible parts from multiple reconstructions can effectively mitigate errors in individual reconstructions and reduce sensitivity to view selection. 
Therefore, we use the combined reconstruction instead of single-view reconstruction to train a unified pose estimator for all 13 objects. After training the low-level layers for 50k steps with learning rate of 2e-4, we train all network layers for 200k steps and then reduce the learning rate to 1e-4 and train 200k steps. 
As shown in Table~\ref{tab:cad_free_lm}, our method significantly outperforms recent CAD-free baselines in terms of the ADD(-S) (0.1d) metric, which demonstrates the capability of our method to handle the more challenging CAD-free setting.

\subsection{Discussion}

\textbf{Qualitative Results.}
In Figure~\ref{fig:qual_bowl},
the predictions of previous methods result in incorrect orientations or physical intersections between the bowl and the sugar box.
In contrast, 
our visibility-aware method estimate accurate orientation and translation (Figure~\ref{fig:qual_bowl}~(e)).
We also visualize the keypoint localization results in Figure~\ref{fig:qual_bowl}~(f), where CheckerPose~\cite{lian2023checkerpose} generates dispersed results with significant outliers.

\textbf{Runtime Analysis.} In Table~\ref{tab:runtime}, we report the running speed of the methods that use Progressive-X~\cite{barath2019progressive} as a PnP solver, on a desktop with an Intel 2.30GHz CPU and an NVIDIA TITAN RTX GPU. The results show clear speed advantage of our method in addition to achieving state-of-the-art accuracies.

\textbf{Non-Convex Objects.} 
For computational efficiency, we approximate internal visibility by Eq.~\ref{eq:visib_inner}. 
In Table~\ref{tab:non_convex}, we analyze the accuracy of the labels on LM, which contains 13 non-convex objects. 
For each object, we uniformly sample 8 views and use ray tracing to compute the ground truth binary visibility labels, which is computationally expensive. For most objects, the accuracy exceeds 80\%. We also examine the accuracy of the keypoints selected by the PPR-based algorithm and observe that PPR effectively mitigates labeling error introduced by the approximation in Eq.~\ref{eq:visib_inner}.

\section{CONCLUSIONS AND FUTURE WORK}

We propose a novel visibility-aware keypoint-based method, named VAPO, for instance-level 6DoF object pose estimation with and without 3D CAD models. 
From object-level annotations, we generate binary keypoint visibility labels as well as real-valued visibility-aware importance, for both asymmetric and symmetric objects. The extensive experiments on LM, LM-O and YCB-V datasets demonstrate that our method significantly improves object pose estimation. 
In future work, we will explore efficient ways to generate visibility labels when object-level annotations are unavailable.
We will also incorporate our visibility-aware importance into zero-shot pose estimation to enhance the visual and geometric embeddings of novel CAD models.







\bibliographystyle{IEEEtran}
\bibliography{main}

\end{document}